\pdfoutput=1

\documentclass[11pt]{article}

\usepackage[final]{acl}

\usepackage{times}
\usepackage{pifont}

\usepackage{latexsym}

\usepackage[T1]{fontenc}

\usepackage[utf8]{inputenc}

\usepackage{microtype}

\usepackage{inconsolata}

\usepackage{graphicx}

\usepackage{amsmath}
\usepackage{amssymb}
\usepackage{multirow}
\usepackage{arydshln}
\usepackage{graphicx}
\usepackage{wrapfig}
\usepackage{colortbl}
\usepackage{booktabs}
\usepackage{makecell}

\usepackage{diagbox}
\usepackage{float}

\RequirePackage{xspace}
\makeatletter
\DeclareRobustCommand\onedot{\futurelet\@let@token\@onedot}
\def\@onedot{\ifx\@let@token.\else.\null\fi\xspace}

\title{Chat-Driven Text Generation and Interaction for Person Retrieval}

\author{
    \textbf{Zequn Xie\textsuperscript{1}},
    \textbf{Chuxin Wang\textsuperscript{1}},
    \textbf{Sihang Cai\textsuperscript{1}},\\
    \textbf{Yeqiang Wang\textsuperscript{2}},
    \textbf{Shulei Wang\textsuperscript{1}},
    \textbf{Tao Jin\textsuperscript{1} \thanks{Corresponding Author.}} \\
    \\
    \normalsize \textsuperscript{1}Zhejiang University \\
    \normalsize \textsuperscript{2} Northwest A\&F University \\
    \normalsize \\ 
    \normalsize Correspondence: zqxie@zju.edu.cn
}

\begin{document}
\maketitle

\begin{abstract}
Text-based person search (TBPS) enables the retrieval of person images from large-scale databases using natural language descriptions, offering critical value in surveillance applications. However, a major challenge lies in the labor-intensive process of obtaining high-quality textual annotations, which limits scalability and practical deployment. To address this, we introduce two complementary modules: \textbf{Multi-Turn Text Generation (MTG)} and \textbf{Multi-Turn Text Interaction (MTI)}. MTG generates rich pseudo-labels through simulated dialogues with MLLMs, producing fine-grained and diverse visual descriptions without manual supervision. MTI refines user queries at inference time through dynamic, dialogue-based reasoning, enabling the system to interpret and resolve vague, incomplete, or ambiguous descriptions—characteristics often seen in real-world search scenarios. Together, MTG and MTI form a unified and annotation-free framework that significantly improves retrieval accuracy, robustness, and usability. Extensive evaluations demonstrate that our method achieves competitive or superior results while eliminating the need for manual captions, paving the way for scalable and practical deployment of TBPS systems.
\end{abstract}

\section{Introduction}

Text-based person search (TBPS) aims to retrieve images of a target individual from large-scale galleries using natural language descriptions~\cite{li2017person}. It lies at the intersection of image-text retrieval~\cite{lei2022loopitr, sun2021lightningdot, miech2021thinking} and image-based person re-identification (Re-ID)~\cite{he2021transreid, luo2019bag, wang2022nformer}, offering a flexible alternative to visual queries. Text queries are more accessible and often provide richer semantic cues about identity, enabling applications ranging from personal photo organization to public security and surveillance.

\begin{figure*}[t]
    \centering
    \includegraphics[width=1 \linewidth]{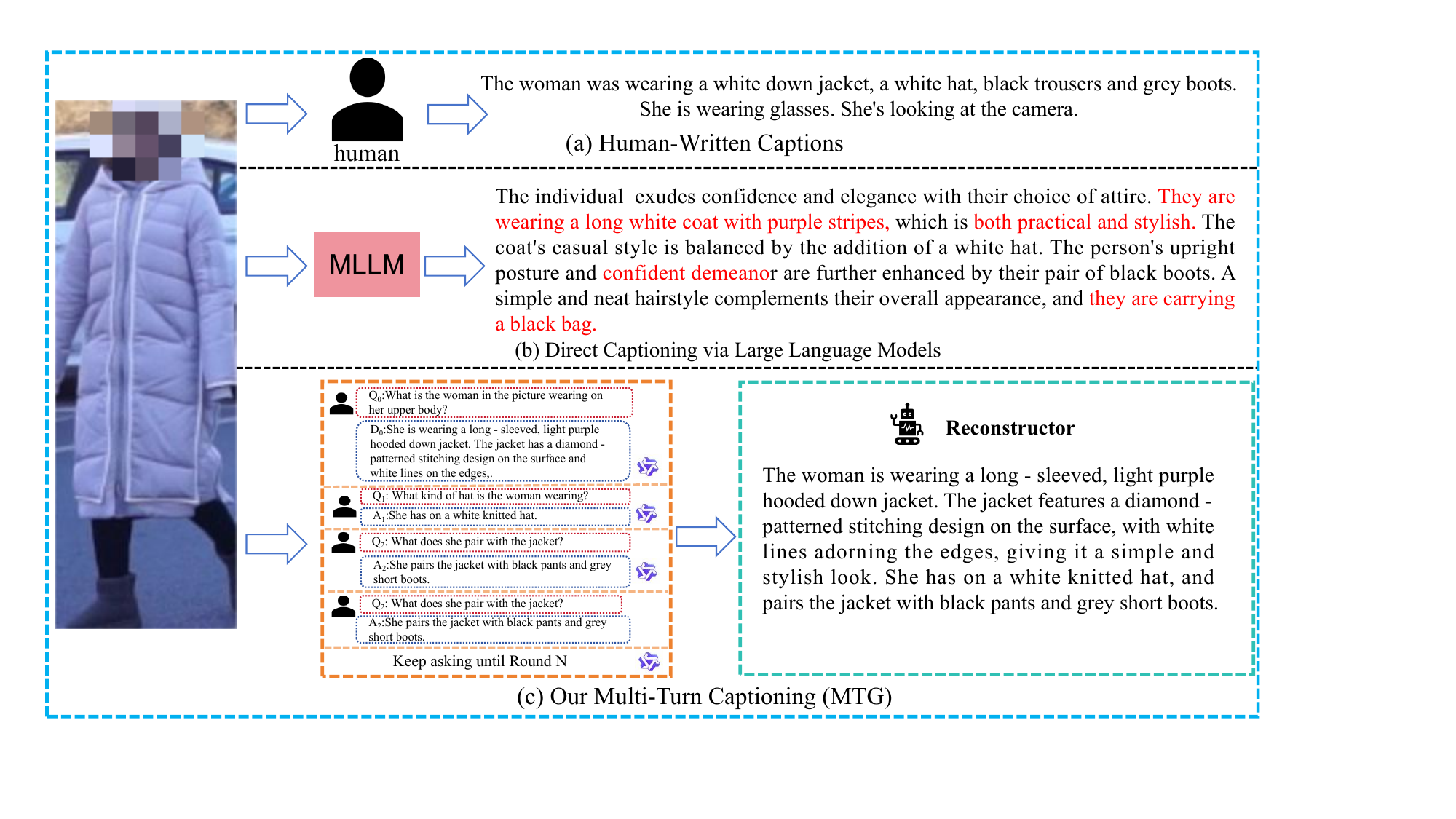}
\caption{Comparison of person description strategies. (a) Human-written captions are concise but often lack compositional depth and attribute coverage. (b) Direct captioning with large language models (LLMs) generates descriptions in a single forward pass, but often suffers from hallucinations or omissions—particularly in capturing fine-grained visual details such as clothing, accessories, or scene context. (c) Our proposed multi-turn strategy simulates an interactive dialogue with the MLLM, progressively enriching descriptions through targeted Q\&A, yielding more expressive, accurate, and human-aligned captions.}

    \label{fig:1}
\end{figure*}

Since the seminal introduction of CUHK-PEDES~\cite{li2017person}, TBPS has made substantial progress, largely driven by advances in cross-modal representation learning that align visual and textual modalities in a shared embedding space~\cite{radford2021learning}. However, despite these technical developments, one fundamental bottleneck remains: the reliance on high-quality textual annotations. While visual data can be easily acquired from surveillance footage, generating accurate and semantically rich descriptions is labor-intensive, expensive, and inherently unscalable. 

Automated captioning methods provide a partial solution, but often suffer from semantic drift, repetitive phrasing, and hallucinated content~\cite{kolouju2025good4cir}, leading to vague or misleading labels (see Figure~\ref{fig:1}). This limitation motivates a central research question: \textit{Can TBPS be achieved effectively without depending on manually crafted descriptions?}

\begin{figure*}[t]
    \centering
    \includegraphics[width=1\textwidth]{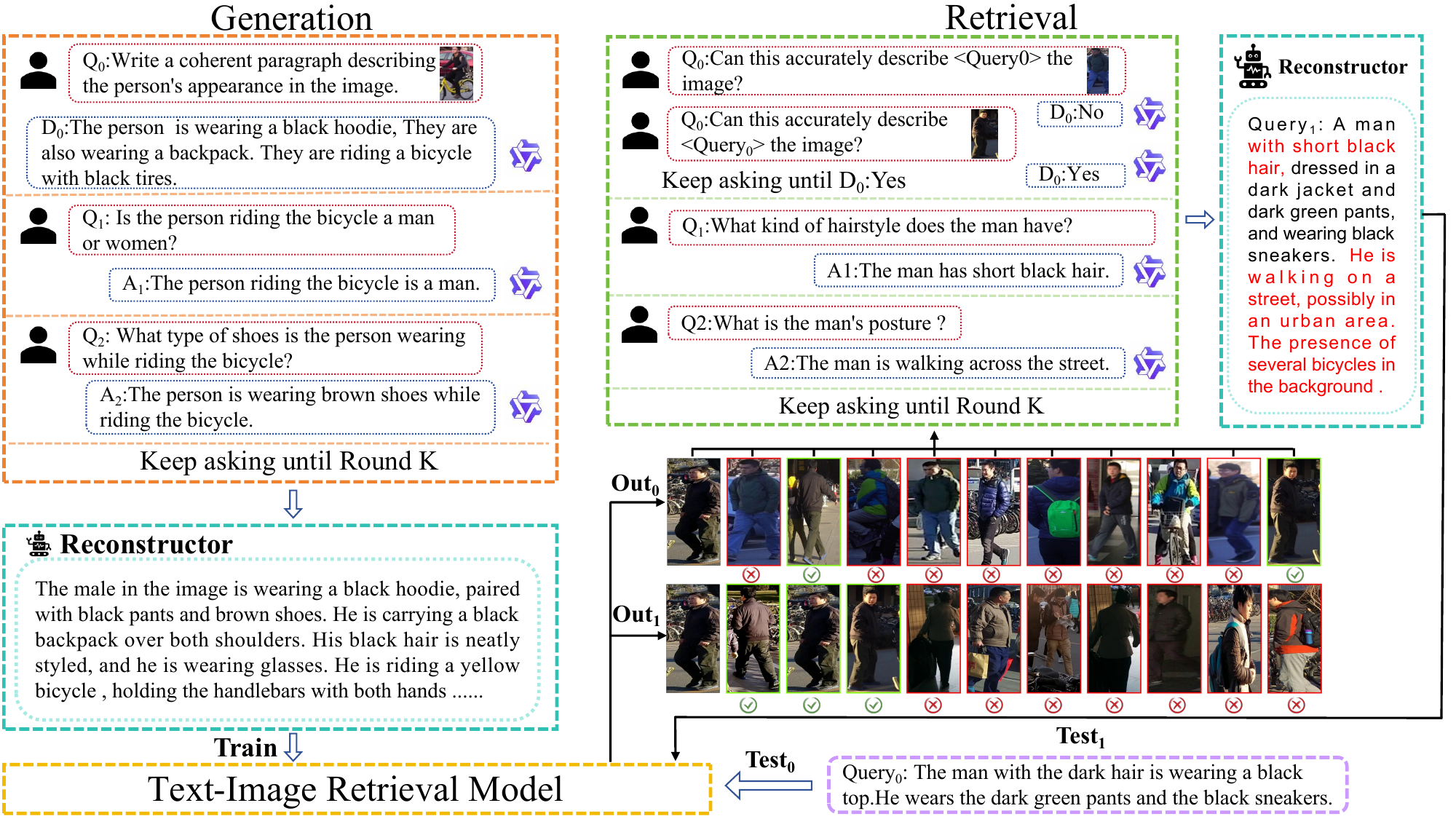}
    \caption{Overview of the proposed \textbf{CTGI} framework for text-based person search. The framework consists of two stages: (1) \textbf{Training-time generation}: MTG simulates multi-turn dialogue to iteratively enrich captions, while a reconstructor synthesizes pseudo-labels using structured prompts; and (2) \textbf{Inference-time retrieval}: MTI refines user queries through MLLM-driven Q\&A, enhancing alignment between the query and candidate images for improved re-ranking.}
    \label{fig:2}
\end{figure*}

To address this challenge, we propose \textbf{Chat-Driven Text Generation and Interaction} (\textbf{CTGI}), a unified and annotation-free framework that bridges the supervision gap via multimodal dialogue. \textbf{CTGI} consists of two synergistic modules: \textbf{Multi-Turn Text Generation} (\textbf{MTG}), which provides training supervision, and \textbf{Multi-Turn Text Interaction} (\textbf{MTI}), which refines queries at inference time (see Figure~\ref{fig:2}).

The \textbf{MTG} module simulates multi-turn conversations with an MLLM to generate rich pseudo-labels. Starting from a baseline caption, it iteratively refines the description using a series of attribute-targeted prompts that mimic human dialogue. This process leads to semantically dense, diverse, and fine-grained annotations that far exceed the quality of single-turn captioning. To accommodate these longer descriptions, we extend CLIP’s default 77-token input limit by applying positional embedding stretching—retaining the first 20 learned positions and interpolating the remaining embeddings to support up to 248 tokens without retraining the model.

The \textbf{MTI} module operates during inference to refine under-specified user queries through MLLM-driven dialogue. It begins by identifying a candidate anchor image and then generates targeted questions to extract missing or ambiguous attributes. The responses are aggregated into a refined query that is better aligned with the target image. MTI also incorporates filtering mechanisms to avoid redundancy and maintain efficiency. As a plug-and-play module, MTI can be easily deployed with various pretrained vision-language retrieval models with minimal adaptation cost.

\textbf{Our key contributions are as follows:}
\begin{itemize}
    \item We propose \textbf{CTGI}, a novel chat-driven framework for TBPS that eliminates the need for manual annotations by unifying pseudo-caption generation and interactive query refinement.
    \item We develop \textbf{MTG}, a multi-turn captioning module that generates rich, attribute-aware pseudo-labels through iterative dialogue, and supports long-text encoding via positional embedding extension.
    \item We introduce \textbf{MTI}, a dynamic inference module that refines natural language queries via MLLM-guided interaction, enhancing alignment between user input and visual content for more accurate retrieval.
\end{itemize}

\section{Related Work}

\subsection{Text-Based Person Search (TBPS)} has progressed significantly since the release of CUHK-PEDES~\cite{li2017person}. Early efforts focused on embedding visual and textual data into a shared space, evolving from global alignment~\cite{zheng2020dual, farooq2020convolutional} to fine-grained matching~\cite{chen2018improvingdeep, chen2022tipcb, suo2022simple}, often enhanced by pose cues~\cite{jing2020pose}, part-level features~\cite{wang2020vitaa}, or semantic knowledge~\cite{loper2002nltk}. In parallel, representation learning approaches aimed to extract modality-invariant features by addressing background clutter~\cite{zhu2021dssl}, color sensitivity~\cite{wu2021lapscore}, and multi-scale fusion~\cite{shao2022learning,wang2024omnibindlargescaleomnimultimodal}. Recently, large-scale pretrained models like CLIP~\cite{radford2021learning} have enabled strong generalization for cross-modal retrieval with minimal tuning~\cite{cvpr23crossmodal, han2021text, wei2023calibrating}, with IRRA~\cite{cvpr23crossmodal} and DURA~\cite{xie2025dynamic} improving alignment via multimodal interaction.
Despite these advances, most TBPS methods still depend on costly human-annotated text, limiting scalability. Weakly supervised~\cite{zhao2021weakly} and synthetic labeling~\cite{yang2023towards, tan2024harnessing} offer partial relief but struggle with vague or conversational queries.

To overcome this limitation, we propose a new TBPS paradigm—\textbf{CTGI}—which eliminates the reliance on manual annotations and significantly enhances retrieval through multi-turn dialogue with MLLMs. In contrast to earlier interactive retrieval systems~\cite{guo2018dialog,lee2024interactive}, which often require task-specific data or model retraining, CTGI supports open-ended queries centered on behavioral attributes and dynamically refines both pseudo-labels during training and user queries during inference. By leveraging MLLMs as plug-and-play agents, CTGI achieves a robust, scalable, and annotation-free TBPS framework.

\subsection*{Vision-Language Pre-training and Multimodal Large Language Models}

The landscape of multimodal research has been fundamentally reshaped by the success of Vision-Language Pre-training. This approach aligns visual and textual data, enabling remarkable zero-shot transfer capabilities that allow models to perform a wide range of downstream tasks, such as text-based image retrieval, without task-specific fine-tuning.
Building on this foundation, recent efforts have advanced MLLMs. Key research directions include enhancing their reasoning abilities~\cite{guo2025observer1unlockingreasoningabilities}, improving underlying representations for better cross-modal generalization~\cite{huang2025opensetcrossmodalgeneralization}, and utilizing prompt-based learning to steer model behavior. Studies have explored techniques such as efficient prompting for adaptation~\cite{guo2025efficient} and dynamic prompt calibration for continual retrieval tasks~\cite{jin2024calibrating}.
Our proposed \textbf{CTGI} framework is directly inspired by these advancements. We leverage the descriptive and conversational power of MLLMs to automate the traditionally manual processes of annotation generation (MTG) and query clarification (MTI). This creates an interactive, annotation-free system for person search.

\section{Methodology}

In this section, we briefly outline a \textbf{C}hat-Driven \textbf{T}ext \textbf{G}eneration and \textbf{I}nteraction (\textbf{CTGI}) model for person retrieval. The CTGI model framework consists of two main modules: (1) The Multi-Turn Text Generation (\textbf{MTG}) module, which uses a multimodal large language model to generate detailed textual descriptions for given person images through an interactive Q\&A dialogue; and (2) The Multi-Turn Text Interaction (\textbf{MTI}) module, which is used in an inference-time pipeline that refines the textual query by leveraging visual context from retrieved images and then performs re-ranking. The overall framework is illustrated in Figure \ref{fig:2}.

\subsection{Multi-Turn Text Generation}
\label{sec:MTG}

The Multi-Turn Text Generation module generates a comprehensive pseudo-label for each person image $I$ by iteratively querying a multimodal large language model for fine-grained details. This process is initiated with an initial captioning prompt designed to elicit a general description. Given an image $I$, we can use the MLLM with a prompt $P_{\text{init}}$, \textit{e.g.}, \textit{``Describe the person in the image,''} yielding an initial static caption $T_s$:
\begin{equation}
    T_s = MLLM \bigl(I, P_{\text{init}}\bigr),
    \label{Eq-1}
\end{equation}
However, $T_s$ provides only a simple, basic textual description and often overlooks distinctive attributes. To capture more distinctive attributes of a person, the QA-guided refinement rounds method provides a more detailed textual description improvement strategy. Specifically, in each round $i$, the model generates an answer $a_i$ that aligns with the image content based on a specific question $q_i$, \textit{e.g.}, 



\begin{quote}
\textit{$q_i$: Is the person riding the bicycle?}\\
\textit{$a_i$: Yes, the person is riding the bicycle.}
\end{quote}

\noindent After $N$ rounds of QA operations, we obtain all preceding QA results \( \{(q_i, a_i)\}_{i=1}^{N} \) concatenated together to obtain the enriched caption \( T_e \):
\begin{equation}
    T_e = MLLM\bigl([a_1, a_2, ..., a_N]\bigr),
    \label{Eq-2}
\end{equation}
Compared to $T_s$, $T_e$ provides more fine-grained attributes for the given person image, \textit{e.g.}, colors, clothing details, and physical features, which greatly enhance the textual description.

It is important to note that due to the presence of similar questions in the question list, this may lead to repetitive answers. To remove the redundant descriptions, we use the MLLM again and reconstruct $T_e$ by incorporating $T_s$:
\begin{equation}
 T_e = MLLM\bigl(T_e \mid T_s, p\bigr),
 \label{Eq-3}
\end{equation}
where $p$ denotes the input prompt to the MLLM, \textit{e.g., ``Rephrase the description using all the above information."}
Compared to $T_e$ in Eq. \eqref{Eq-2}, $T_e$ in 
Eq. \eqref{Eq-3} provides a more concise and effective textual description, rather than increasing the quantity of image-related details. Meanwhile, compared to $T_s$ 
in Eq.\eqref{Eq-1}, $T_e$ contains more details extracted during the MLLM Q\&A process, and better aligns with human attention to core image information.

\subsection{More Text Positional Embeddings}

CLIP’s original 77-token limit, imposed by its fixed-length absolute positional embeddings, restricts its ability to process long and detailed text—a critical limitation for tasks such as Text-Based Person Search (TBPS). To address this, we adopt a knowledge-preserving \textit{positional embedding stretching} technique that extends the model’s input capacity while maintaining compatibility with pretrained weights.

Following Long-CLIP~\cite{zhang2024long}and FineLIP\cite{asokan2025finelip}, we preserve the first 20 learned positional embeddings, which are empirically the most well-trained, and interpolate the remaining positions (21–77) to reach a new input length of 248 tokens by applying a 4$\times$ stretching factor.

Let $PE(pos)$ denote the original positional embedding at position $pos \in [1, 77]$. We construct the stretched embedding $PE^*(pos)$ for the extended range $pos \in [1, 248]$ as:

\begin{equation}
\scriptsize
PE^*(pos) = 
\begin{cases} 
PE(pos), & \text{for } pos \leq 20 \\
(1 - \alpha) \cdot PE\left(\left\lfloor \frac{pos}{\lambda_2} \right\rfloor \right) \\
\quad + \alpha \cdot PE\left(\left\lceil \frac{pos}{\lambda_2} \right\rceil \right), & \text{for } 21 \leq pos \leq 77
\end{cases}
\end{equation}

Here, $\lambda = \frac{248 - 20}{77 - 20} \approx 4$ is the interpolation factor, and $\alpha$ is the fractional part of $\frac{pos - 20}{\lambda}$. This ensures smooth interpolation while preserving pretrained embeddings for the initial positions.

Inspired by LiT~\cite{zhai2022lit}, this approach avoids reinitialization or retraining, and allows CLIP to encode longer, semantically rich descriptions generated by the MTG module. Empirical results in Table \ref{tab:4} confirm that this strategy enhances retrieval performance without sacrificing alignment learned during pretraining.

\subsection{Multi-Turn Text Interaction (MTI)}

MTI operates during inference to resolve underspecified or vague user queries through multi-turn interaction.

\textbf{Step 1: Anchor Identification.}  
Given a user query $q$, the system retrieves top-$K$ candidates $\{\hat{v}_1, ..., \hat{v}_K\}$ using similarity score $S_{q,v}$. For each $\hat{v}_k$, the MLLM is prompted to judge alignment with $q$. The first affirmative response identifies the anchor $\bar{v}$. If no match is found within $K$ attempts, no refinement is applied.

\textbf{Step 2: Interactive Refinement.}  
With anchor $\bar{v}$, MTI generates a diagnostic question set $\{c_i\}$ focused on missing attributes. Responses are obtained via visual Q\&A:

\begin{equation}
    r_{\bar{v}} = \text{MLLM}(T_{\text{vqa}}(\{c_i\}, \bar{v}))
\end{equation}

The final query $\hat{q}$ is synthesized using a template prompt to merge $r_{\bar{v}}$ and $q$:

\begin{equation}
    \hat{q} = \text{MLLM}(T_{\text{aggr}}(r_{\bar{v}}, q))
\end{equation}

\textbf{Step 3: Re-ranking.}  
The final similarity is computed as:

\begin{equation}
    \hat{S}_{q,v} = \lambda S_{q,v} + (1 - \lambda) S_{\hat{q}, v}
\end{equation}

with $\hat{S}_{q, \bar{v}} = 1$ to promote anchor matching. Early stopping is triggered when $\hat{v}_1$ surpasses threshold $\xi = 0.85$.

\subsection{Reconstructor}
\label{sec:reconstructor}

The \textbf{Reconstructor} plays a pivotal role in transforming fragmented outputs from multi-turn Q\&A into coherent and high-quality descriptions. It is deployed in both training and inference pipelines to enhance the effectiveness of CTGI without requiring any manual annotations or dataset-specific tuning.
To ensure the quality of generated descriptions during training, MTG maintains a dynamic question pool and discards Q\&A pairs that exhibit low semantic relevance or redundant information. This filtering helps avoid overlong or repetitive captions.

For synthesis, the Reconstructor leverages the \textbf{GPT-4o API} to convert structured Q\&A logs into fluent and semantically rich pseudo-captions. These refined captions serve as supervision signals for training downstream retrieval models.

In the inference stage, the Reconstructor also contributes to query refinement within MTI. A set of curated diagnostic templates is used to identify typical ambiguities. These templates help elicit missing attributes without introducing generic or noisy questions. The responses are then aggregated into a revised query that is semantically aligned with the visual anchor.

This unified design ensures that CTGI can support both training-time pseudo-label generation and test-time query refinement effectively—without reliance on human-written descriptions or task-specific engineering.

\section{Experiments}

To validate the efficacy and robustness of our proposed framework, we conduct a comprehensive set of experiments. Our evaluation strategy begins by re-annotating three public datasets, generating enriched textual descriptions that provide greater semantic depth and diversity than the original captions. We then systematically compare retrieval models trained on these generated pseudo-labels against baseline models trained on the original annotations. To assess the framework's generalizability, we integrate our method into standard TBPS pipelines to evaluate its impact. Finally, a series of in-depth ablation studies and visual analyses are performed to comprehensively dissect the contributions of each component and better understand the method's overall effectiveness.

\begin{table*}[t]
\caption{Performance on CUHK-PEDES . *: trained with LLaVA-1.5 captions. The best and second-best results are in \textbf{bold} and \underline{underline}, respectively.}
  \label{tab:1}
\begin{center}
\resizebox{1 \textwidth}{0.3\textheight}{ 
\begin{tabular}{l|lcc|cccc}\toprule[1pt] 
\textbf{Methods}            & \textbf{Ref.}       &  \textbf{Image Enc.     } & \textbf{Text Enc.}      & \textbf{ R-1}  & \textbf{R-5 }  & \textbf{ R-10}  &  \textbf {mAP }     \\\midrule

\multicolumn{8}{@{}l}{\textbf{Fully Supervised}} \\
\midrule

TIMAM~\cite{sarafianos2019adversarial}              & ICCV'19   & RN101      & BERT         & 54.51  & 77.56  & 79.27   & -        \\
ViTAA~\cite{wang2020vitaa}              & ECCV'20   & RN50       & LSTM         & 54.92  & 75.18  & 82.90   & 51.60   \\
NAFS~\cite{gao2021contextual}               & arXiv'21  & RN50       & BERT         & 59.36  & 79.13  & 86.00   & 54.07 \\
DSSL~\cite{zhu2021dssl}               & ACMMM'21     & RN50       & BERT         & 59.98  & 80.41  & 87.56   & -        \\
SSAN~\cite{ding2021semantically}               & arXiv'21  & RN50       & LSTM         & 61.37  & 80.15  & 86.73   &  -       \\
Lapscore~\cite{wu2021lapscore}           & ICCV'21   & RN50       & BERT         & 63.40   & -      & 87.80   & -       \\
ISANet~\cite{yan2022image}             & arXiv'22  & RN50       & LSTM         & 63.92  & 82.15  & 87.69   & -         \\

SAF~\cite{li2022learning}                & ICASSP'22 & ViT-Base   & BERT         & 64.13  & 82.62  & 88.40   & -         \\

DCEL~\cite{qin2022deepdecl}&ACMMM'22&CLIP-ViT& CLIP-Xformer &   71.36&  {88.11}&  {92.48}& 64.25\\
IVT~\cite{shu2022see}                & ECCV'22  & ViT-Base   & BERT         & 65.59  & 83.11  & 89.21   & -        \\
CFine~\cite{yan2022clip}              & TIP'23  & CLIP-ViT   & BERT         & 69.57  & 85.93  & 91.15   & -      \\
IRRA~\cite{jiang2023cross}               & CVPR'23   & CLIP-ViT   & CLIP-Xformer & 73.38  & 89.93  & 93.71   & 66.13 \\
BiLMa~\cite{fujii2023bilma}    & ICCV'23   & CLIP-ViT   & CLIP-Xformer  &74.03& 89.59& 93.62 &66.57\\ 

PBSL~\cite{shen2023pedestrian}&ACMMM'23 & RN50&BERT&65.32 &83.81 &89.26&-\\
BEAT~\cite{ma2023beat}&ACMMM'23&RN101&BERT& 65.61& 83.45&89.54&-\\ 
LCR$^2$S~\cite{yan2023learning}&ACMMM'23&RN50&TextCNN&67.36& 84.19& 89.62& 59.24\\
DCEL~\cite{li2023dcel}&ACMMM'23&CLIP-ViT& CLIP-Xformer &   75.02&  {90.89}&  {94.52}& -\\
UniPT~\cite{shao2023unified}&ICCV'23&CLIP-ViT& CLIP-Xformer &68.50& 84.67 &-&-\\ 
TBPS~\cite{cao2024empirical}&AAAI'24&CLIP-ViT& CLIP-Xformer&73.54& 88.19 &92.35& 65.38 \\
{RDE}~\cite{qin2024noisy} & CVPR'24 & CLIP-ViT& CLIP-Xformer & {75.94} &{90.14}& {94.12} &{67.56}\\  

CFAM ~\cite{zuo2024ufinebench}&CVPR'24&CLIP-ViT& CLIP-Xformer&75.60& 90.53&-&67.27\\
MGRL ~\cite{lv2024mgrl}&ICASSP'24&CLIP-ViT& CLIP-Xformer&73.91& 90.68&-& 67.28\\
OCDL ~\cite{li2025object}&ICASSP'25&CLIP-ViT& CLIP-Xformer&75.10& 89.43&-& 68.18\\
\midrule

\multicolumn{8}{@{}l}{\textbf{Unsupervised}} \\
\midrule

IRRA* ~\cite{li2025exploring} & CVPR'23 & CLIP-ViT& CLIP-Xformer& 32.94 & 54.37 & 64.67 & 30.87 \\
BLIP* ~\cite{li2025exploring} & ICML'22 &BLIP-ViT & BLIP-Xformer& 51.41 & 71.41 & 78.76 & 44.73 \\

GTR~\cite{bai2023text}    & MM'23 &BLIP-ViT & BLIP-Xformer& 47.53 & 68.23 &75.91 & 42.91 \\

{MUMA}~\cite{li2025exploring}& AAAI'25 &BLIP-ViT & BLIP-Xformer & {59.52} & {77.79} & - & {52.75} \\

Ours (with IRRA)
 & EMNLP'25 & CLIP-ViT& CLIP-Xformer &  \underline{63.53} &\underline{80.25}  &\underline{87.84} &\underline{52.37} \\
Ours (with RDE)
 & EMNLP'25 & CLIP-ViT& CLIP-Xformer&\textbf{67.82} & \textbf{85.45}&\textbf{90.63 }&\textbf{55.14}

  \\
\bottomrule[1pt]
\end{tabular}}

\end{center}
\end{table*}

\subsection{Datasets and Performance Measurements}
\textbf{Datasets.} We evaluate our approach using three Text-based Person
Retrieval datasets: CUHK-PEDES~\cite{12}, ICFG-PEDES~\cite{23}, and RSTPReid~\cite{32}. Our training solely utilizes image data, devoid of any dependency on manually annotated text data. During the testing phase, captions from the dataset are leveraged for re-
trieval.

\textbf{Evaluation Metrics.} Following standard practice, we evaluate using Rank-k (k=1,5,10), mean Average Precision (mAP). Higher values indicate better retrieval performance.

\subsection{Implementation Details}

We evaluate \textbf{CTGI} using two strong TBPS baselines: \textbf{IRRA}~\cite{IRRA} and \textbf{RDE}~\cite{qin2024noisy}, both built on \textbf{CLIP-ViT/B-16}~\cite{radford2021learning}. For multimodal reasoning, we adopt \textbf{Qwen2-VL-7B}~\cite{Qwen2VL} as the core MLLM, while the \textbf{Reconstructor} leverages the \textbf{OpenAI GPT-4o API}~\cite{openai2023gpt4o} for pseudo-caption synthesis.

All models follow the original training setups of IRRA and RDE. Input images are resized to $384 \times 128$, and standard augmentations (flip, crop, erase) are applied. To support longer text, we extend CLIP’s 77-token limit to \textbf{248 tokens} by preserving the first 20 positional embeddings and interpolating the rest 4$\times$, following~\cite{zhai2022lit}. The learning rate is set to $1\times10^{-5}$ (with 5 warmup epochs from $1\times10^{-6}$), and $5\times10^{-5}$ for randomly initialized layers. Cosine decay is used throughout 60 training epochs.

During training, the \textbf{MTG} module runs \textbf{6 Q\&A rounds} per image to generate dense pseudo-labels. For inference, \textbf{MTI} examines the top \textbf{$K=20$} retrieval candidates, and early exits if the top-1 similarity exceeds $\xi = 0.85$ and is confirmed by the MLLM. Final retrieval scores are fused via weighted re-ranking. All experiments are conducted on \textbf{2$\times$ NVIDIA RTX 4090 GPUs} with generation temperature fixed at \textbf{0.01} for stability.

\subsection{Comparison with the State-of-the-Art}

We evaluate the effectiveness of our proposed CTGI framework on three widely used benchmark datasets for text-based person search, comparing against both unsupervised and fully supervised state-of-the-art methods. Our framework is instantiated with two variants, \textit{Our+IRRA} and \textit{Our+RDE}, which employ different retrieval backbones while sharing the same underlying CTGI components.

\textbf{CUHK-PEDES:} As reported in Table~\ref{tab:1}, under the unsupervised setting, our \textit{Our+RDE} achieves a Rank-1 of 67.82\% and mAP of 55.14\%, substantially outperforming the strongest unsupervised baseline MUMA, which obtains 59.52\% and 52.75\% respectively. Notably, \textit{Our+IRRA} also surpasses MUMA by a clear margin, demonstrating the strong efficacy of CTGI in generating informative pseudo-labels and improving retrieval without manual annotations. Compared with fully supervised methods, our results approach competitive levels, surpassing several mid-tier supervised models and narrowing the gap to the top performers.

\textbf{ICFG-PEDES:} Table~\ref{tab:2} shows that our framework maintains state-of-the-art performance in the unsupervised category with a Rank-1 of 56.16\% and mAP of 32.40\% for \textit{Our+RDE}, exceeding the best supervised methods in some metrics. This highlights CTGI’s robustness and generalization ability across datasets with different granularity and annotation styles. The improvements over other unsupervised baselines such as BLIP and GTR further confirm the superiority of our approach.

\textbf{RSTPReid:} As shown in Table~\ref{tab:3}, on the RSTPReid dataset, \textit{Our+RDE} achieves a Rank-1 of 66.35\% and mAP of 51.51\%, outperforming the second-best unsupervised method MUMA by approximately 12\% in Rank-1 and over 11\% in mAP. Moreover, our method exceeds the performance of several fully supervised models, including CFine, illustrating the strong competitiveness and scalability of CTGI without reliance on any manual annotations.

Across all datasets, our CTGI framework demonstrates a consistent and significant improvement over existing unsupervised methods, closing the gap towards fully supervised performance. These results validate the effectiveness of leveraging multimodal large language models for pseudo-label generation and interactive query refinement, enabling robust and scalable text-based person search in practical scenarios.

\begin{table}
\centering
\caption{Performance on ICFG-PEDES. *: trained with LLaVA-1.5 captions.The best and second-best results are in \textbf{bold} and \underline{underline}, respectively.}
\resizebox{1\linewidth}{!}{
\begin{tabular}{l|cccc}
\toprule[1pt]
\textbf{Method} & \textbf{R@1} & \textbf{R@5} & \textbf{R@10} & \textbf{mAP} \\
\midrule
\multicolumn{5}{l}{\textbf{Fully Supervised}} \\
\midrule
Dual Path~\cite{zheng2020dual}         & 38.99 & 59.44 & 68.41 & - \\
CMPM/C~\cite{zhang2018deep}            & 43.51 & 65.44 & 74.26 & - \\
ViTAA~\cite{wang2020vitaa}             & 50.98 & 68.79 & 75.78 & - \\
SSAN~\cite{ding2021semantically}       & 54.23 & 72.63 & 79.53 & - \\
IVT~\cite{shu2022see}                  & 56.04 & 73.60 & 80.22 & - \\
ISANet~\cite{yan2022image}             & 57.73 & 75.42 & 81.72 & - \\
CFine~\cite{yan2022clip}               & 60.83 & 76.55 & 82.42 & - \\
IRRA~\cite{jiang2023cross}             & 63.46 & 80.25 & 85.82 & 38.06 \\
BiLMa~\cite{fujii2023bilma}            & 63.83 & 80.15 & 85.74 & 38.26 \\
PBSL~\cite{shen2023pedestrian}         & 57.84 & 75.46 & 82.15 & - \\
BEAT~\cite{ma2023beat}                 & 58.25 & 75.92 & 81.96 & - \\
LCR$^2$S~\cite{yan2023learning}        & 57.93 & 76.08 & 82.40 & 38.21 \\
DCEL~\cite{li2023dcel}                 & 64.88 & 81.34 & 86.72 & - \\
UniPT~\cite{shao2023unified}           & 60.09 & 76.19 & -     & - \\
TBPS~\cite{cao2024empirical}           & 65.05 & 80.34 & 85.47 & 39.83 \\
CFAM~\cite{zuo2024ufinebench}          & 65.38 & 81.17 & -     & 39.42 \\
MGRL~\cite{lv2024mgrl}                 & {67.28} & 63.87 & -     & {82.34} \\
OCDL~\cite{li2025object}               & 64.53 & 80.23 & -     & 40.76 \\
\midrule
\multicolumn{5}{l}{\textbf{Unsupervised}} \\
\midrule
IRRA* ~\cite{li2025exploring}         & 21.23 & 37.37 & 46.04     & 11.47 \\
BLIP* ~\cite{li2025exploring}          & 31.58 & 52.03 &  61.73    & 13.20 \\
GTR ~\cite{bai2023text}                 & 28.25 & 45.21 &  53.51     & 13.82 \\
MUMA ~\cite{li2025exploring}         & {38.11} & {56.01} & 63.96 & {19.02} \\
Ours (with IRRA)
                           &  \underline{48.76} &  \underline{67.38}  &  \underline{   74.66}    & \underline{ 27.42}    \\

Ours (with RDE)
&\textbf{56.16}  & \textbf{73.18}   &\textbf {79.42 }     &\textbf{32.40}   \\
\bottomrule[1pt]

\end{tabular}}
\label{tab:2}
\end{table}

\begin{table}[h]
\centering 
\caption{Performance on RSTPReid. *: trained with LLaVA-1.5 captions.The best and second-best results are in \textbf{bold} and \underline{underline}, respectively.}
\resizebox{1\linewidth}{!}{
\begin{tabular}{l|ccccc}
\toprule[1pt] 
\textbf{Methods} & \textbf{R-1}& \textbf{R-5} & \textbf{R-10}& \textbf{mAP}   \\
\midrule
\multicolumn{5}{@{}l}{\textbf{Fully Supervised}} \\
\midrule
DSSL~\cite{zhu2021dssl}   & 39.05& 62.60  & 73.95& -          \\
SSAN~\cite{ding2021semantically}   & 43.50& 67.80  & 77.15& -     \\
LBUL~\cite{wang2022look}   & 45.55& 68.20  & 77.85& -         \\
IVT~\cite{shu2022see}    & 46.70& 70.00  & 78.80& -          \\
CFine~\cite{yan2022clip}  & 50.55& 72.50  & 81.60& -          \\
IRRA~\cite{jiang2023cross}  & 60.20 & 81.30  & 88.20 & 47.17 \\ 
BiLMA~\cite{fujii2023bilma}&61.20& 81.50& 88.80 &48.51\\
PBSL~\cite{shen2023pedestrian}&47.80& 71.40& 79.90&-\\
BEAT~\cite{ma2023beat}&48.10 &73.10 &81.30&-\\
LCR$^2$S~\cite{yan2023learning}&54.95& 76.65& 84.70 &40.92\\
DCEL~\cite{li2023dcel}& 61.35& 83.95 &{90.45}&-\\
TBPS~\cite{cao2024empirical} &61.95& 83.55& 88.75& 48.26\\
CFAM ~\cite{zuo2024ufinebench}&62.45& 83.55&-&49.50\\

OCDL ~\cite{li2025object}&61.60&82.35&-& 49.77\\

\midrule
\multicolumn{5}{@{}l}{\textbf{Unsupervised}} \\
\midrule
IRRA* ~\cite{li2025exploring} & 37.60 & 60.65 &72.30 & 27.42 & - \\
BLIP* ~\cite{li2025exploring} &  44.45 & 67.70 &77.25 & 33.73 & - \\

GTR ~\cite{bai2023text}   & 45.60 & 70.35 &79.95 & 33.30  \\
{MUMA} ~\cite{li2025exploring}  & 54.35 & 76.05 & 83.65  & 40.50 \\
Ours (with IRRA)
  & \underline{ 64.20}&  \underline{83.55} &  \underline{90.30} &  \underline{49.66}   \\
Ours (with RDE)
& \textbf{ 66.35} &  \textbf{85.50}&  \textbf{ 91.25} & \textbf{51.51}  \\

\bottomrule[1pt]
\end{tabular}}
  \label{tab:3}
\end{table}

\subsection{Ablation Study}

We conduct ablation experiments on the RSTPReid dataset to systematically analyze the individual and combined effects of MTG and MTI. When employed separately, MTG enhances retrieval by generating detailed and semantically rich pseudo-labels, resulting in notable improvements in Rank-1 accuracy and mAP over the baseline. For instance, with the IRRA backbone, MTG alone achieves a Rank-1 of 52.30\%, indicating its strong ability to provide effective training supervision through enriched textual descriptions.

Similarly, MTI, which refines user queries at inference time via multi-turn dialogue, independently boosts performance by improving the semantic alignment between queries and visual features. This is reflected by an increased Rank-1 accuracy of 55.50\% with IRRA, highlighting MTI's effectiveness in mitigating ambiguity in free-form textual queries.

Importantly, the integration of MTG and MTI yields complementary benefits, producing the highest gains across all metrics. Combined, they achieve Rank-1 accuracies of 64.20\% and 66.35\% with IRRA and RDE backbones respectively, alongside corresponding mAP improvements. These results confirm that the synergy between richer pseudo-label generation and dynamic query refinement substantially advances cross-modal retrieval performance and robustness.

\begin{table}[h]
\centering
\caption{Ablation study on the RSTPReid dataset. MTG: Multi-Turn Text Generation, MTI: Multi-Turn Text Interaction, PES: Positional Embedding Stretching.}
\label{tab:4}
\resizebox{0.5\textwidth}{!}{
    \begin{tabular}{l|cc|cccc}
    \toprule
    \textbf{Method} & \textbf{MTG} & \textbf{MTI} & \textbf{Rank-1} & \textbf{Rank-5} & \textbf{Rank-10} & \textbf{mAP} \\
    \midrule
    Ours (with IRRA)
                & \checkmark &            & 52.30 & 74.65 & 84.05 & 40.03 \\
    Ours (with IRRA)
               &           & \checkmark  & 55.50 & 77.50 & 86.55 & 44.87 \\
    Ours (with IRRA)
                & \checkmark & \checkmark  & 64.20 & 83.55 & 90.30 & 48.03 \\
    Ours (with IRRA) (w/o PES)        & \checkmark & \checkmark  & 63.00 & 82.65 & 88.80 & 47.60 \\
    \midrule
    Ours (with RDE)                     & \checkmark &            & 60.55 & 79.85 & 86.30 & 44.98 \\
    Ours (with RDE)                   &           & \checkmark  & 62.55 & 82.85 & 89.00 & 46.43 \\
    Ours (with RDE)                   & \checkmark & \checkmark  & 66.35 & 85.50 & 91.25 & 49.66 \\
    Ours (with RDE)   (w/o PES)         & \checkmark & \checkmark  & 65.75 & 84.05 & 90.60 & 49.60 \\
    \bottomrule
    \end{tabular}
}
\end{table}

\newcommand{\yes}{\textcolor{green!60!black}{\checkmark}}
\newcommand{\no}{\textcolor{red!70!black}{\ding{55}}}

\begin{figure*}[t]
    \centering
    \includegraphics[width=0.94\linewidth]{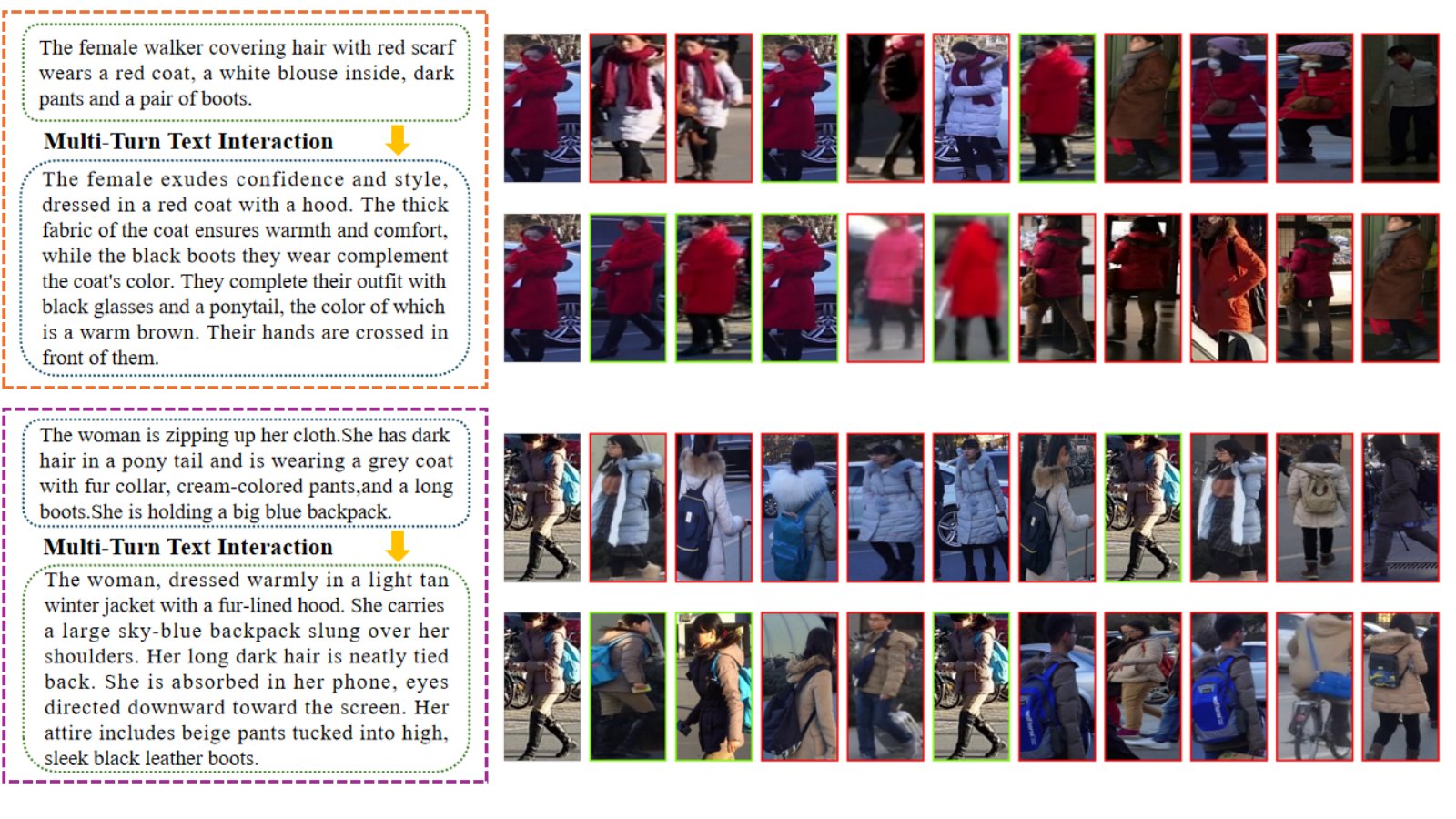}
    \caption{Top-10 retrieval results on the RSTPReid dataset. The first column is the ground-truth image. The first row shows retrieval results using IRRA; the second row shows results after applying IRRA with MTI. Refined queries generated by multi-turn interaction are shown alongside each example. Green borders indicate correct matches.}
    
    \label{fig:3}
\end{figure*}

\subsection{Visualization of Retrieval Results}
To evaluate the effectiveness of MTI, we conducted controlled experiments with a fixed operation cycle. Figure~\ref{fig:3} visualizes the top-10 retrieval results before and after applying MTI. Notably, the retrieval model is trained solely on pseudo-captions generated by the MTG module, without any manual annotations. Due to the incomplete alignment between initial queries and ground-truth test captions, retrieval without MTI often yields suboptimal results. In contrast, MTI dynamically refines the query through interactive optimization, enabling more accurate and robust ranking performance.

\section{Conclusion}

In this work, we introduced \textbf{CTGI} (Chat-Driven Text Generation and Interaction), a unified and annotation-free framework for Text-Based Person Search (TBPS) that removes the dependency on manually crafted textual descriptions. CTGI integrates two synergistic modules: \textbf{Multi-Turn Text Generation (MTG)} for training supervision and \textbf{Multi-Turn Text Interaction (MTI)} for inference-time refinement. Together, they leverage the expressive capabilities of Multimodal Large Language Models (MLLMs) to generate rich pseudo-labels and iteratively enhance user queries via natural language dialogue. Extensive experiments across multiple TBPS benchmarks show that CTGI achieves competitive or superior performance compared to fully supervised methods, while seamlessly adapting to existing retrieval pipelines. Ablation studies and qualitative visualizations further underscore the value of multi-turn interaction and MLLM-guided refinement in improving cross-modal alignment and retrieval robustness.

\section*{Limitations}

While \textbf{CTGI} demonstrates strong performance without manual annotations, several challenges remain. First, pseudo-labels generated by MTG may contain semantic noise or redundancy. Although robust retrieval backbones like RDE are designed for noisy environments and thus benefit more from such supervision, other models without inherent noise-filtering may be more vulnerable to degraded performance. Second, MTI introduces additional inference overhead due to multi-turn interactions with MLLMs. Even with early stopping and anchor validation, this can limit deployment in latency-sensitive applications. Third, both MTG and MTI rely on the generalization ability of the underlying MLLM, which may yield suboptimal results in unfamiliar domains or when handling fine-grained attributes. Future work could address these issues through uncertainty-aware label filtering, more efficient MLLMs, and domain-adaptive interaction strategies.

\section*{Acknowledgments}
\begin{sloppypar}
This work was supported by the ``Pioneer'' and ``Leading Goose'' R\&D Program of Zhejiang under (Grant No. 2025C02110), Public Welfare Research Program of Ningbo under (Grant No. 2024S062), and Yongjiang Talent Project of Ningbo under (Grant No. 2024A-161-G).    
\end{sloppypar}

\bibliography{main}

\newpage

\newpage

\appendix

\end{document}